\documentclass[sigconf]{acmart}

\makeatletter                   
\def\mdseries@tt{m}             
\makeatother                    
\usepackage{hyperref}
\usepackage[plain]{fancyref}
\usepackage[draft=true]{minted} 
\usepackage{color}
\hypersetup{
    colorlinks=true,
    linkcolor=blue,
    filecolor=red,      
    urlcolor=magenta,
    breaklinks=true,            
}
\usepackage{breakurl}           
\usepackage{amsthm}
\usepackage{subcaption}

\def\BibTeX{{\rm B\kern-.05em{\sc i\kern-.025em b}\kern-.08emT\kern-.1667em\lower.7ex\hbox{E}\kern-.125emX}}
    
\copyrightyear{2019}
\acmYear{2019}
\setcopyright{acmlicensed}
\acmConference[SigChi '19]{Where is the Human? Bridging the Gap Between AI and HCI, CHI Workshop}{Glasgow}{UK}

\begin{document}
\sloppy                         
%
\title{Fairness Sample Complexity and the Case for Human Intervention}

%

\author{Ananth Balashankar}
\affiliation{
\institution{New York University, Google Research}
\city{New York}
\country{USA}
}
\email{ananth@nyu.edu}

\author{Alyssa Lees}
\affiliation{
\institution{Google Research}
\city{New York}
\country{USA}
}
\email{alyssalees@google.com}

%
%
\begin{abstract}
With the aim of building machine learning systems that incorporate standards of fairness and accountability, we explore explicit subgroup sample complexity bounds. The work is motivated by the observation that classifier predictions for real world datasets often demonstrate drastically different metrics, such as accuracy, when subdivided by specific \textit{sensitive variable} subgroups.  The reasons for these discrepancies are varied and not limited to the influence of mitigating variables, institutional bias, underlying population distributions as well as sampling bias. Among the numerous definitions of fairness that exist, we argue that at a minimum, principled ML practices should ensure that classification predictions are able to mirror the underlying sub-population distributions.  However, as the number of sensitive variables increase, populations meeting at the intersectionality of these variables may simply not exist or may not be large enough to provide accurate samples for classification. In these increasingly likely scenarios, we make the case for human intervention and applying situational and individual definitions of fairness. In this paper we present lower bounds of subgroup sample complexity for metric-fair learning based on the theory of Probably Approximately Metric Fair Learning.
We demonstrate that for a classifier to approach a definition of fairness in terms of specific sensitive variables, adequate subgroup population samples need to exist and the model dimensionality has to be aligned with subgroup population distributions.  In cases where this is not feasible, we propose an approach using individual fairness definitions for achieving alignment.  We look at two commonly explored UCI datasets under this lens and suggest human interventions for data collection for specific subgroups to achieve approximate individual fairness for linear hypotheses.
 
\end{abstract}

\maketitle

\section{Introduction}
In recent discussions of ethical ML algorithms, evaluating fairness has been frequently predicated on defining constraints based on specific sensitive variables, such as race or gender.  These variables should \textbf{not} demonstrate conditionally discriminative behavior while learning classification targets.  If care is not taken in the construction of a ML model, works such as \cite{Zhao2017MenAL} and \cite{EnsignFNSV17} have shown that inequalities in underlying data distributions can be amplified in the predicted output, leading to runaway feedback loops.

Recent works \cite{Kearns2018PreventingFG} have argued that examining the intersectionality of multiple sensitive variables is crucial for establishing coherent standards of fairness. However, real world data sub-populations often display varying underlying sampling distributions, bias and noise. 
We argue that principles towards \textit{fair} ML should encourage subgroups to perform individually optimally in a classification task and at a minimum be able to reflect their true underlying population distributions.  

As the number of sensitive variables increase, the intersectional subgroup populations tend to  decrease in size.  In fact, it is not unlikely to have a dataset with a subgroup population of one or zero. In these scenarios, it is evident  that any classifier which achieves non-trivial accuracy can never be \textit{fair}. In order to overcome this tradeoff, we look to the rich literature of ``individual fairness'' which  defines fairness with respect to a similarity metric between two individuals and enforces that individuals who are similar are treated similarly, within an error bound \cite{Dwork2012FairnessTA, Dwork2019FairnessUC, Kearns2018PreventingFG}. We find this definition to be useful in measuring the complexity of a class of machine learning models that attempt to achieve approximate \textit{individual fairness} on a dataset.

Using such a model complexity measure, we can compute the minimum number of samples required in each sub-group to learn approximately ``individually fair'' models with high probability \cite{pacf}. Sample complexity provides us a lower bound on the number of samples required from sub-group populations before we can make any claims of learning a \textit{fair} model. This methodology provides us a principled way of thinking about the guarantees provided by any algorithm which claims to learn a fair model, similar to how confidence intervals are used for regression coefficients. If the guarantees aren't feasible, principled human interventions need to be undertaken to better estimate the confidence of the underlying fair machine learning system.

We have been motivated by the insight that many fairness problems in existing classification tasks for specific subpopulations can been remedied by increased data collection, subject to ethical considerations \cite{Buolamwini2018GenderSI, Chen2018WhyIM}. As such, we suggest that in the spirit of achieving approximately metric-fair learning, sub-population sample size lower bounds need to be aligned with the actual subgroup sample sizes in the dataset. In this paper, we highlight the discrepancies in model and sample complexities for sub-populations and the disalignment of the actual sizes with these complexities in two popular UCI datasets. 
Given that sampling bias leads to unintended consequences in ML classifiers, we show the need to gather new data of specific subgroups in prevalent fairness datasets to avoid perpetrating such conditions induced from biased or insufficient sampling. 

\section{Related work}
Defining required sample complexity for ML algorithms to achieve low generalization error has been studied extensively in ML theory literature \cite{mohribook, vcBound}. Probably Approximately Correct (PAC)  Learning defines upper bounds on generalization error related to a sample's empirical error and complexity of the hypotheses and datasets \cite{Valiant:1984:TL:1968.1972}. More recently, \cite{behnam, arora} focused on deriving tighter bounds on deep neural networks in the regime of overparameterization based on compression. Related to this paper's focus on overcoming dimensionality constraints for multiple subgroups, \cite{blum} explores the problem of learning a single objective using multiple actors simultaneously, which provides an exponential improvement in sample complexity as opposed to no collaboration. 
Recent works by \cite{Kleinberg2018SimplicityCI} also highlight that simple hypotheses cause inequity by design. Our work extends on this, providing a methodology for quantifying sampling bounds for smaller informational datasets and highlights the pitfalls of basing a decision making system on a simple linear hypotheses in some circumstances. 

\section{Limits of Fair Sensitive Variable Subgroup Sampling}
The use of explicit sensitive variables in real-world scenarios is sometimes a hard constraint.  One example is legislation enforcing fairness around disparate impact \cite{disparateimpact, Feldman:2015:CRD:2783258.2783311}.

In simplified examples, exploring the intersectionality of sensitive variables may be appropriate. For example, in this paper we explore two gender and two race subgroups in the evaluation of the UCI Adult dataset. In our simplified example of subgroups this translates into four separate populations.  It is conceivable that in a real-world application, the intersection of gender and race subgroups could extend into \textbf{many} different populations.  As the intersectionality of subgroups grow, it is likely that a subgroup's sample population will be insufficient.  In the case of the UCI German dataset used in this paper, marriage status, with five possible values, is treated as a sensitive variable along with gender. However, in the dataset there were \textbf{no} samples containing both the attributes of Female and Single. 

It is obvious that there is no model of any complexity that can achieve full fairness with respect to the sensitive variables in this case. Increased data collection is obviously a first step in these situation.  However, what if the true subgroup population simply does not exist?  In these scenarios, the case for alternate definitions of individual fairness should be explored. 

Similarly, in order to satisfy fairness constraints on these increasing intersectionalities of populations, it is necessary that we revisit the framework we use to measure the efficacy of the modeling choices. Despite the impossibility results of achieving fairness in the extreme case of subgroup sized one, there is still a need to highlight cases where simple (linear) models are inadequately applied in datasets with complex underlying subgroup distributions \cite{propublica, chouldechova}. The ability to objectively quantify the required model and sample complexity to satisfy these fairness constraints will guide the choices made by practitioners and ML researchers.

\section{Approximate Fairness Sample Complexity Bounds}
In order to overcome the impossibility results surrounding individual fairness, the definition of Probably Approximately Metric-Fair learning was first introduced in \cite{pacf}, which formalized the guarantees of fairness expected. 
\begin{definition}
Given a data  distribution $D$, hypothesis $h$ trained on a sample $S$ of size $m$ drawn in an i.i.d manner, the hypothesis $h$ is said to be $(\alpha, \gamma)$ approximately metric-fair with respect to a similarity metric $d$ between two inputs $x$ and $x'$ which discards any variations in sensitive attributes, if 
\[
Pr_{x, x' \sim D}[|h(x) - h(x')| > d(x, x') + \gamma] \leq \alpha
\]
\end{definition}

This definition is closely linked with Rademacher model complexity measure  \cite{Bartlett:2003:RGC:944919.944944}, which measures the capacity of a class of models (hypotheses) to fit the given dataset with any random error vector for the dataset. We present a relevant result from \cite{pacf} below which gives us the fairness generalization guarantee for a class of models for a given dataset.

\begin{theorem}
If the Rademacher complexity for a hypothesis class $H$, $R_{m}(H) = \frac{r}{\sqrt{m}}$   then, for all $\delta, \epsilon_{\alpha}, \epsilon_{\gamma} \in (0,1)$, there exists a sample complexity
\[m = O(\frac{r^{2} \cdotp ln(\frac{1}{\delta})}{\epsilon_{\alpha}^2 \cdotp \epsilon_{\gamma}^2}) \]
such that with probability (1 - $\delta$) over the sample $S$, $ h \in H$, if $h$ is $(\alpha, \gamma)$ approximately metric-fair on sample S, then $h$ is $(\alpha + \epsilon_{\alpha}, \gamma + \epsilon_{\gamma})$ approximately metric-fair on the underlying distribution $D$.
\end{theorem}

This means that if we found a hypothesis $h$  which is probably approximately fair over $m$ samples, then that hypothesis would also be probably approximately fair over the true (unsampled) data distribution, albeit with different parameters.

\begin{lemma}
Rademacher complexity bounds for linear hypotheses $H = \{x \rightarrow w \cdotp x\}$, is given by $R_m(H) \leq \frac{R \cdotp \phi}{\sqrt{m}}$ where $R = max ||x||_{2}$ and $\phi = max||w||_{2}$ \cite{mohribook}
\end{lemma}

Combining Lemma 4.2 with Theorem 4.1 , we get that the sample complexity for approximately metric-fairness guarantees for linear hypotheses 
\begin{align}
m = O(\frac{R^{2} \cdotp \phi^{2} \cdotp ln(\frac{1}{\delta})}{\epsilon_{\alpha}^2 \cdotp \epsilon_{\gamma}^2})
\end{align}
This complexity applies to all $h \in H$.  However, if an exponential-time probably approximate metric-fair learning algorithm, that evaluates empirical error for all hypotheses and selects the $h$ that minimizes it, is applied, it results in a smaller sample complexity
\begin{align}
    m = O(\frac{R^{2} \cdotp \phi^{2} \cdotp ln(\frac{1}{\delta})}{min(\epsilon_{\alpha}, \epsilon_{\gamma})^2})
\end{align}
This is equivalent to the information theoretical sample complexity in \cite{pacf}. Since, the set of linear hypotheses is the one applicable to the widely used linear and logistic regression models applied to the UCI datasets and other crucial interpretable ML problems, this forms the basis of our critical analysis. More complex models like the deep neural networks, are currently out of scope for this analysis, though non-linear models may be more appropriate for the designated task. 

\section{Subgroup Sample Complexity}
In constructing subgroup sample bounds, the above sample complexity measure is applied on various subgroups $\mathcal{P}(G)$ of a dataset. The decision to measure sample complexities per subgroup instead of the overall dataset is motivated by the necessity to highlight the discrepancies in the underlying distributions. Achieving approximate fair learning on the entire dataset is only possible when the sample complexities of the sub-groups are met. If not possible, human interventions in the data collection process by either correcting sampling bias or even collecting new data for subgroups not well represented is advised.

We define membership of an instance to an intersectionality of groups $G = <g_1, g_2...g_n>$, for different values of n ($< |X|$) sensitive variables. Calculating the sample complexity for each of the subgroups provides a quantitative measure of the differences between subgroups for a given hypotheses class $H$. Significant discrepancies indicate differences in the underlying distributions, subgroup sampling methodologies or true labeling functions for these subgroups. In some instances, the sample complexity bound on the overall population may suggest if further sampling should be performed before attempting to learn a single model applied on all subgroups indiscriminately. These complexity metrics provide a clear directive of the order of new samples that needs to be collected before we can make any claims of probably approximately metric-fairness on the hypothesis class. The guarantees associated with it need to match the desired operating surface of $\delta, \epsilon_{\alpha}, \epsilon_{\gamma}$ before we can utilize $H$ for safe decision making.  Also, since these measures are distribution and hypothesis specific, it provides us a much tighter bound than purely combinatorial measures which work for any distribution and learning algorithms.

\subsection{Efficiency bounds}
While approximate fair learning ensures that similar data points are classified approximately similarly, there may also be requirements on the minimum accuracy achieved by each of the sub-groups. While this might be at odds with the fairness requirement, it still forms the basis of recent bias mitigating ML literature that rely on constrained optimization \cite{pmlr-v81-menon18a}. 
We can obtain a lower bound on the sample complexity to learn a $(\epsilon, \delta)$ PAC objective function in a collaborative manner \cite{blum} among k subgroups. Specifically, there exists a hypothesis h chosen from a class $H$ of VC dimension d, such that it has generalization error less than $\epsilon$ on all the k subgroups' distributions $\{D_1, D_2, .. D_k\}$ with high probability $1-\delta$. Current literature also provides algorithms for both a centralized and personalized setting, where either a single function is learned for all k subgroups, or each subgroup can have its own function \cite{blum}. In the centralized setting, the sample complexity to learn a PAC algorithm which achieves error of less than $\epsilon$ in all k subgroups is lower bounded by
\begin{align}
m = O(\frac{ln^2(k)}{\epsilon}((d+k)ln(\frac{1}{\epsilon}) + kln(\frac{1}{\delta})) 
\end{align}
For the personalized setting, the sample complexity is $O(ln(k))$ factor smaller than the centralized sample complexity. For uniform convergence lower bounds where $\epsilon, \delta \in (0,0.1)$, there exists a PAC learnable hypothesis class of VC dimension d for $m \geq dk(1-\delta)/(4\epsilon)$. This provides a minimum bound for achieving a threshold of error rates across all subgroups, which is compatible with the minimum efficiency fairness intent. 

\section{Evaluation}
We evaluated the empirical Radamacher and associated sample complexities for two datasets, the UCI Adult Census dataset and the UCI German Credit dataset \cite{uci} over a linear hypothesis class.

In the UCI Adult Census dataset, income bracket prediction is based on a set of individual categorical and numerical attributes like age, education, occupation, etc. In this paper, we specify race and gender, as \textit{sensitive variables} that should not inform our model in accurately predicting the income bracket. The computed Rademacher complexities, broken down for the intersection of all subgroup combinations for this dataset shown in Table \ref{tab:rad_adult}, demonstrate that some subgroups are harder to learn than others for a linear hypothesis.

In order to calculate the discrepancy in sub-group sample complexity for the UCI Adult dataset, approximate metric fairness lower bounds for the underlying distributions were ordered by sub-groups for a given $\delta, \epsilon_{\alpha}, \epsilon_{\gamma}$.
A condensed table is shown in  Table \ref{tab:sam_adult}. We used one-hot encoding for categorical features and hence the number of dimensions for a linear separating hyperplane d, is 108. 
With k = 4 subgroups, the lower bound for PAC learning is $\frac{432(1 - \delta)}{\epsilon}$. To calculate the associated complexity, we trained logistic regression models for each subgroup with 10 fold cross validation to choose regularization weights in \cite{scikit-learn}. We calculated the maximum norms of the normalized inputs (R) and the coefficients ($\phi$). Note that since the lower bound sample complexity for PAC-fair learning is denoted in the big $O$ notation (Equation 1), we can only estimate the ordering or ranking of the sample complexities for the subgroups in Table \ref{tab:sam_adult} (higher numbered rank has higher complexity values). 
The ordering of the actual subgroup sample sizes ($4 > 2 > 3 > 1$) reveals that new samples are needed to match the desired sample complexity ordering ($3 > 2 > 4 > 1$). Specifically, more samples for subgroup 3 need to be gathered than for from subgroups 2 or 4 to ensure the ordering of actual sample sizes align with that of the sample complexities.

\begin{table}
\footnotesize
\begin{subtable}{1\linewidth}\centering
{\begin{tabular}{lc}
\hline
\textbf{Subgroup (per gender/race)} & \textbf{$R_m(H)$}\\\hline
1 (Female/Black) & 0.113\\
2 (Female/White) & 1.002\\
3 (Male/Black) & 2.998 \\
4 (Male/White) & 0.649\\\hline
Overall & 0.561\\
\end{tabular}}
\caption{UCI Adult dataset}
\label{tab:rad_adult}
\normalsize
\end{subtable}%

\bigskip
\begin{subtable}{1\linewidth}\centering
\footnotesize
{\begin{tabular}{lc}
\hline
\textbf{Subgroup (per status/sex)} & \textbf{$R_m(H)$}\\\hline
1 (Male/Separated) & 1.776\\
2 (Female/Separated-Married) & 1.051\\
3 (Male/Single) & 0.349 \\
4 (Male/Married) & 0.001\\\hline
Overall & 0.301\\
\end{tabular}}
\caption{UCI German Credit dataset}
\label{tab:rad_german}
\normalsize
\end{subtable}%
\caption{Discrepancy in Rademacher complexity bounds of linear hypotheses for sensitive subgroups}
\label{tab:rad}
\end{table}%

\begin{table}
\footnotesize
\begin{subtable}{1\linewidth}\centering
\resizebox{\columnwidth}{!}{%
{\begin{tabular}{lcc}
\hline
\textbf{Subgroup} & \textbf{Sample Complexity Rank} & \textbf{Actual Sample Size (Rank)}\\\hline
1 & 1 & 2,129 (1)\\
2 & 3 & 8,642 (3)\\
3 & 4 & 2,616 (2)\\
4 & 2 & 19,174 (4)\\\hline
\end{tabular}}}
\caption{UCI Adult dataset}
\label{tab:sam_adult}
\normalsize
\end{subtable}%

\bigskip
\begin{subtable}{1\linewidth}\centering
\footnotesize
\resizebox{\columnwidth}{!}{%
{\begin{tabular}{lcc}
\hline
\textbf{Subgroup} & \textbf{Sample Complexity Rank} & \textbf{Actual Sample Size (Rank)}\\\hline
1 & 3 & 50 (1)\\
2 & 4 & 310 (3)\\
3 & 2 & 548 (4)\\
4 & 1 & 92 (2)\\
\end{tabular}}}
\caption{UCI German Credit dataset}
\label{tab:sam_german}
\normalsize
\end{subtable}
\caption{Comparison of sample complexity ranking for Probably Approximately Metric Fairness with actual subgroup sizes of subgroups}
\label{tab:sam}
\end{table}

In the UCI German Credit dataset, personal financial attributes are present with a target objective of predicting good/bad credit.  We treat 'personal status' and 'sex' as sensitive attributes.  Since there was no data available for all possible sub-combinations of the 5 personal status options, we treat 'personal status and sex' as a compound feature and examine the 4 subgroups existing with membership in the dataset. The lower bound on the PAC learning sample complexity where d=61, k=4 is $\frac{244(1-\delta)}{\epsilon}$. Table \ref{tab:rad_german} and Table \ref{tab:sam_german} show disparity in the order of the actual sample sizes ($3 > 2 > 4 > 1$) as compared to desired sample complexity ($2 > 1 > 3 > 4$). This implies that in the UCI German Credit dataset, more new samples from group 2 than from group 3 should be drawn as prescribed by the sample complexity. Similarly, more samples from subgroup 1 need to be collected than  from subgroup 4 in order to remove any inversion in ranking of complexities and actual subgroup sizes to ensure guarantees of probably approximate metric-fair learning using linear hypotheses.

\section{Conclusion}
We have suggested an empirical methodology to quantify the discrepancy between distributions of sensitive subgroups based on the theory of sample complexity for probably approximately metric-fair learning. The evaluation on two extensively used datasets in the fairness ML literature, highlight the shortcomings of claims that a linear hypotheses can be probably approximately fair for these two populations. We demonstrate the need for further sampling from particular subgroups before such a conclusion can be made.
Future work to extend this analysis to more complex ML models may provide a principled standard for ensuring subpopulation fairness. We argue in the development of an ethical AI framework for policy and decision makers, sufficient subgroup sampling and the corresponding model complexity should be of prime focus. When adequate subpopulation samples are not feasible, human intervention is advisable. 

\bibliographystyle{unsrt}
\bibliography{sample-sigconf}
\end{document}